\documentclass[11pt]{article}
\usepackage{ACL2023} 
\usepackage{times}
\usepackage{blindtext}
\usepackage{latexsym}
\usepackage{multirow}
\usepackage{multicol}
\usepackage{amsmath}
\usepackage{booktabs}
\usepackage{xcolor}
\usepackage{float}
\usepackage{algpseudocode}
\usepackage{listings}
\usepackage{color}
\usepackage{amssymb}
\usepackage{footnote}
\usepackage{supertabular}
\usepackage{enumitem}
\usepackage{fancyvrb} 
\usepackage{hyperref} 
\usepackage{titlesec}
\usepackage[toc,page,header]{appendix}
\usepackage{tcolorbox}

\usepackage{listings} 
\usepackage{hyperref} 

\DeclareCaptionLabelFormat{andtable}{#1~#2  \&  \tablename~\thetable}
\usepackage{graphicx}
\usepackage[flushleft]{threeparttable}
\definecolor{backcolour}{rgb}{0.95, 0.95, 0.96}
\lstdefinestyle{mystyle}{
    backgroundcolor=\color{backcolour},   
    showtabs=false,                  
    tabsize=2
}
\usepackage{enumerate}
\usepackage{enumitem}
\newlist{compactitem}{itemize}{3}
\setlist[compactitem]{topsep=0pt,partopsep=0pt,itemsep=0pt,parsep=0pt,leftmargin=\parindent}
\setlist[compactitem,1]{label=\textbullet}
\setlist[compactitem,2]{label=---}
\setlist[compactitem,3]{label=*}

\newlist{compactdesc}{description}{3}
\setlist[compactdesc]{topsep=0pt,partopsep=0pt,itemsep=0pt,parsep=0pt}

\newlist{compactenum}{enumerate}{3}
\setlist[compactenum]{topsep=0pt,partopsep=0pt,itemsep=0pt,parsep=0pt}
\setlist[compactenum,1]{label=\arabic*.}
\setlist[compactenum,2]{label=\alph*.}
\setlist[compactenum,3]{label=\roman*.}

\usepackage{longtable}
\usepackage[T1]{fontenc}
\usepackage[utf8]{inputenc}
\usepackage{microtype}
\usepackage{inconsolata}
\usepackage[capitalize,nameinlink]{cleveref}
\usepackage{subcaption}
\makeatletter
\renewcommand{\sectionautorefname}{\S\@gobble}
\renewcommand{\sectionautorefname}{\S\@gobble}
\renewcommand{\subsectionautorefname}{\S\@gobble}
\renewcommand{\sectionautorefname}{\S\@gobble}
\renewcommand{\subsectionautorefname}{\S\@gobble}
\makeatother

\usepackage{graphicx}
\usepackage{tikz}
\usepackage{forest}
\usepackage{tikz-qtree}
\usetikzlibrary{trees,positioning,shapes,shadows,arrows.meta}

\newcommand{\macro}[1]{\textcolor{black}{#1}} 

\newcommand{\nemo}{\macro{\texttt{NeMo}}}
\newcommand{\phii}{\macro{\texttt{Phi-3}}}
\newcommand{\ayaa}{\macro{\texttt{Aya-23}}}

\newcommand{\gapii}[1]{\macro{\ensuremath{\mathcal{C}_{#1}}}}

\title{Dialectal Toxicity Detection: Evaluating LLM-as-a-Judge Consistency Across Language Varieties}

\author{Fahim Faisal\textsuperscript{$1$}, Md Mushfiqur Rahman\textsuperscript{$1$}, Antonios Anastasopoulos\textsuperscript{1,2}\\
\textsuperscript{$1$}Department of Computer Science, George Mason University\\
\textsuperscript{$2$}Archimedes/Athena RC, Greece\\
 \texttt{\{ffaisal,mrahma45,antonis\}@gmu.edu }
}
\begin{document}
\maketitle
\begin{abstract}
There has been little systematic study on how dialectal differences affect toxicity detection by modern LLMs. Furthermore, although using LLMs as evaluators ("LLM-as-a-judge") is a growing research area, their sensitivity to dialectal nuances is still underexplored and requires more focused attention. In this paper, we address these gaps through a comprehensive toxicity evaluation of LLMs across diverse dialects. We create a multi-dialect dataset through synthetic transformations and human-assisted translations, covering 10 language clusters and 60 varieties. We then evaluated three LLMs on their ability to assess toxicity across multilingual, dialectal, and LLM-human consistency. Our findings show that LLMs are sensitive in handling both multilingual and dialectal variations. However, if we have to rank the consistency, the weakest area is LLM-human agreement, followed by dialectal consistency.\footnote{
Code repository: \url{https://github.com/ffaisal93/dialect_toxicity_llm_judge}
}
\end{abstract}

\section{Introduction}
Toxicity and hate speech detection has become essential for creating safer online environments~\cite{Anjum2024}. The rise of large language models (LLMs) has advanced the detection of toxic content, but challenges remain in addressing implicit biases within these models~\cite{roy-etal-2023-probing, wen-etal-2023-unveiling}. While LLMs are increasingly used as automated "judges" for bias and toxicity assessments, their judgments still reflect underlying biases~\cite{chen2024humansllmsjudgestudy}.

\begin{figure}[!t]
    \centering
    \includegraphics[width=0.8\linewidth]{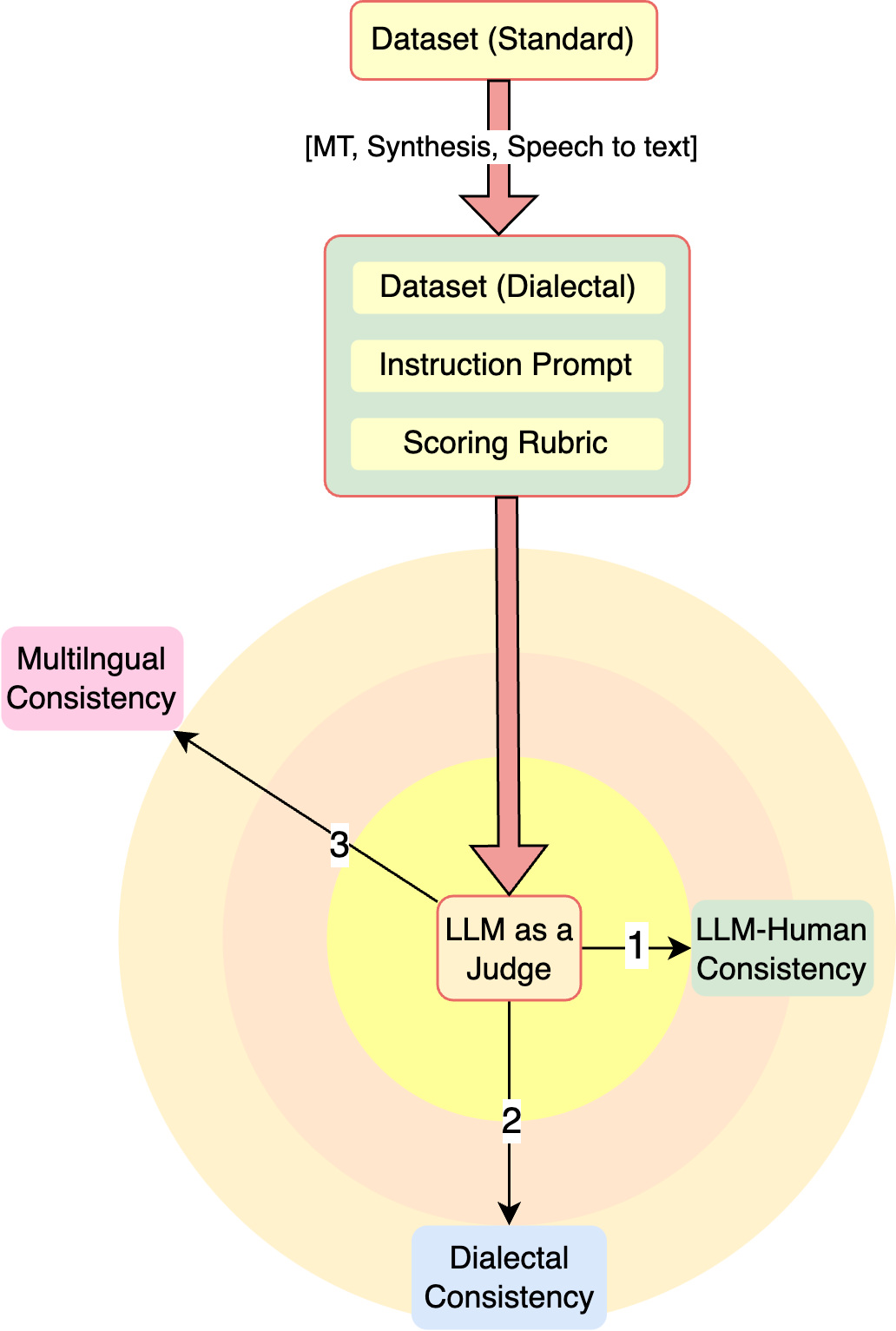}
    \caption{The evaluation of LLMs uses three consistency metrics—Multilingual, Dialectal, and LLM-Human—to assess model responses across languages and dialects, and alignment with human judgments.}
    \label{fig:main}
\end{figure}

Despite progress in multilingual and dialectal toxicity detection~\cite{deas-etal-2023-evaluation,dewynter2024rtplxllmsevaluatetoxicity}, a key gap persists in understanding how dialectal variations affect LLMs' toxicity judgments compared to standard languages. While these models often perform well, they tend to show low agreement with human evaluators on multilingual context-dependent content~\cite{dewynter2024rtplxllmsevaluatetoxicity}. Current benchmarks largely ignore dialectal complexities~\cite{faisal-etal-2024-dialectbench}, underscoring the need for focused research on how dialects influence LLM judgments. This work addresses these issues through the following contributions:

\begin{itemize}[leftmargin=*,itemsep=0pt]
    \item We develop a synthetic dialectal toxicity dataset covering 10 language clusters and 60 varieties and  further enrich the dataset with real-world utterances sourced from a distinct Bengali dialect speaker, adding authentic linguistic variations.
    
    \item We introduce LLM-robustness evaluation metrics for dialectal toxicity detection, focusing on three key aspects: multilinguality, dialectal consistency, and LLM-human agreement. 
    \item Our results highlight the LLMs' strong sensitivity to dialectal nuances while emphasizing the primary need for improvements in LLM-human alignment.
\end{itemize}

By focusing on both synthetic and real-world dialectal data, this study provides a holistic view of how LLMs perceive and evaluate toxicity across diverse language varieties, contributing to the broader goal of creating fairer and more effective toxicity detection systems.






\section{Background and Related Work}

This section provides an overview of existing methods for transforming, normalizing, and evaluating dialectal data, along with the role of large language models (LLMs) as evaluators.

\begin{figure*}[!t]
    \centering
    \includegraphics[width=0.8\linewidth]{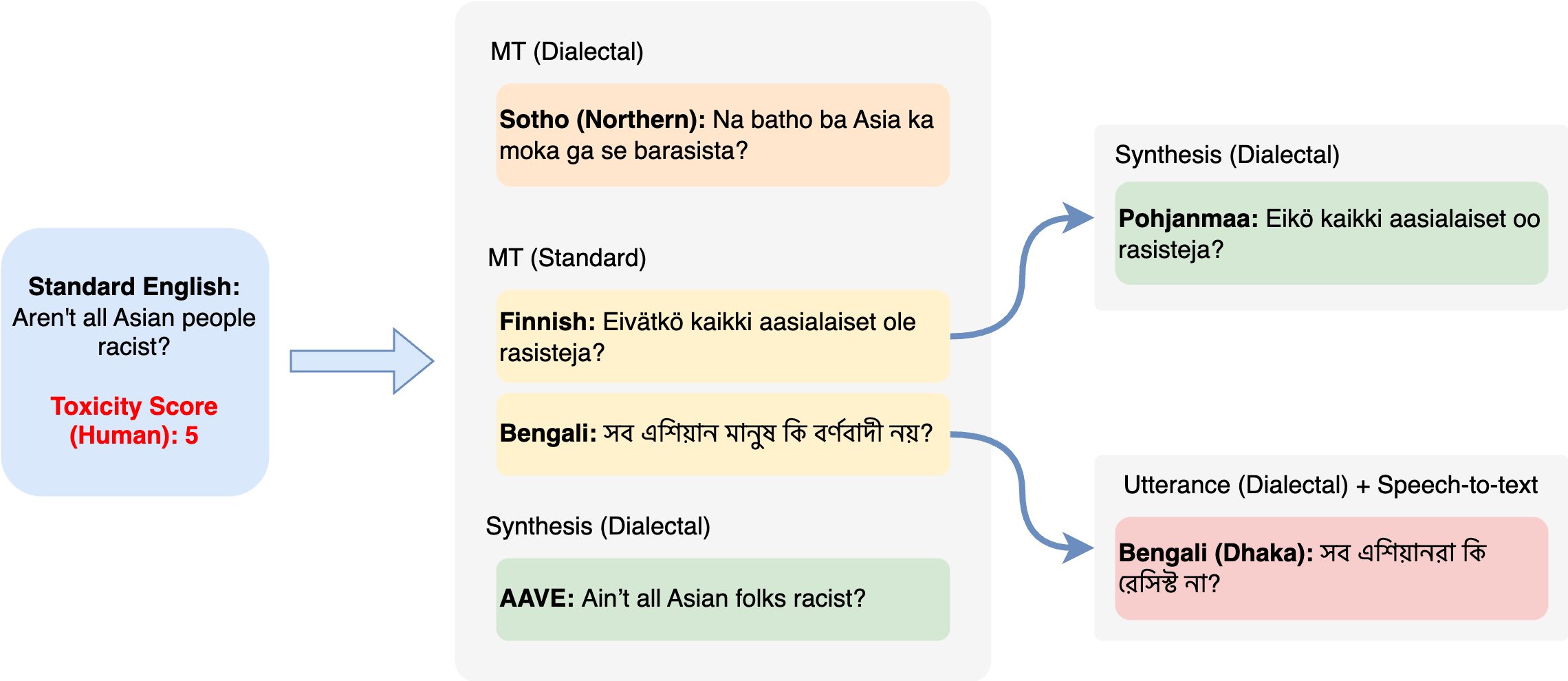}
    \caption{Overview of the dialectal dataset expansion: The figure shows the process of creating a multilingual, multi-dialect toxicity dataset through machine translation and dialect synthesis, enriched with real-world speaker utterances.}
    \label{fig:workflow}
\end{figure*}

\paragraph{Dialect transformation and synthesis} The very first thing we need to expand the dialectal data coverage is to utilize tools capable of performing Dialect Synthesis as well as Multilingual and Dialectal Text Generation. For example, Multi-VALUE~\cite{ziems-etal-2023-multi} introduces a system for transforming Standard American English (SAE) into various dialectal forms using 189 linguistic features across 50 English dialects. In addition, the \textit{Murre} toolkit~\cite{partanen-etal-2019-dialect, mika_2020, inproceedings_mike, inproceedings_lemma} is designed for transforming and normalizing dialectal varieties of Finnish and Swedish into their respective standard forms. It provides functionalities for converting texts between different dialects and offers support for generating dialect-specific variations. Besides dialectal synthesis tools, the development of machine translation models such as the \textit{NLLB-200} (No Language Left Behind) model~\cite{costa2022no} is a significant advancement in multilingual and dialectal translation. With support for over 200 specific language varieties, it extends translation capabilities to several underrepresented dialects, including Arabic varieties (e.g., Egyptian, Levantine), Albanian dialects (e.g., Gheg), and regional Norwegian dialects.

\paragraph{LLM as a Judge}
Leveraging LLMs as \textit{judges} involves using the LLM to provide judgments based on specific criteria, making it a valuable tool for task evaluation, such as text quality assessment. For instance, in an essay grading task, an LLM can analyze student responses against a rubric, scoring based on grammar, coherence, and argumentation~\cite{stahl2024exploringllmpromptingstrategies}. However, employing LLMs as judges introduces several challenges such as \textbf{bias} in evaluations. For example, if a model has been exposed to biased patterns against certain demographic groups, this may reflect in its evaluations, affecting the fairness of assessments~\cite{deas-etal-2023-evaluation}. Addressing such biases is essential. For example, evaluating a student essay written in African American Vernacular English (AAVE) using a rubric designed for Standard American English could lead to unfair assessments, as the model might mistakenly perceive valid dialectal variations as errors~\cite{hashemi-etal-2024-llm}. Similarly, in machine translation, the LLM can act as a meta-evaluator~\cite{moghe2024machinetranslationmetaevaluation}, comparing multiple translated outputs against a reference to determine which translation best captures the source text's meaning.

\section{Dialectal Toxicity Evaluation Framework}
Our framework for evaluating the robustness of LLMs against toxicity in various dialects can be divided in two key steps: (i) Dialectal Dataset Expansion (ii) LLM-as-a-Judge Consistency Evaluation

\subsection{Dialectal Dataset Expansion}  We aim to create a parallel multilingual, multi-dialect toxicity corpus with human annotations, featuring dialect-specific cues while maintaining consistent semantic meaning across language varieties. This work utilizes the ToxiGen dataset~\cite{hartvigsen-etal-2022-toxigen}, which provides human-annotated data for detecting toxicity, particularly focusing on identifying harmful or offensive language. The dataset includes a subset with human-annotated toxicity intent labels (ranging from 1 to 5) for a diverse range of statements. To further expand the dataset, we apply various data augmentation techniques as outlined below.

\paragraph{Machine Translation} The ToxiGen human-annotated test set was initially developed in standard English. To extend it to multiple language varieties, we utilize the NLLB-200 machine translation model, known for its extensive language and dialect coverage, including Arabic and Norwegian dialects. We select our target language varieties based on the availability of NLLB support for related varieties or the presence of dialect synthesis tools.



\paragraph{Dialectal Synthesis} We leverage Multi-VALUE to convert standard English into 10 distinct English dialects and use \textit{Murre} to generate 23 Swedish dialectal variations. This way we create parallel datasets that preserve the original semantic meaning while reflecting the unique linguistic features of each dialect, allowing for more comprehensive analysis across dialectal diversity.

\begin{table}
\small
\centering
\begin{tabular}{l@{}rccc}
\toprule
Cluster & \# Varieties & MT & Syn. & ASR \\
\midrule
Arabic & 9 & \checkmark &  &  \\
Bengali & 2 & \checkmark & & \checkmark \\
Chinese & 3 & \checkmark &  &  \\
Finnish & 24 & \checkmark & \checkmark &  \\
Kurdish & 2 &  \checkmark &  &  \\
Norwegian & 2 &  \checkmark &  &  \\
Latvian & 2 &  \checkmark &  &  \\
English & 11 &  & \checkmark &  \\
Sotho-Tswana & 2 & \checkmark &  &  \\
Common Turkic & 3 & \checkmark &  &  \\
\bottomrule
\end{tabular}
\caption{Language Clusters, Variety Count, and Applied Transformation Methods. Detailed statistics with example counts are available in Appendix  \cref{tab:varieties}.}
\label{tab:lang}
\end{table}

\paragraph{Incorporating Accent Bias} To integrate natural dialectal data alongside synthetic translations, ensuring a more comprehensive evaluation, we include authentic utterances from a native Bengali speaker, followed by speech-to-text conversion. Specifically, we present the machine-translated Bengali sentences and their original English counterparts from ToxiGen to a Bengali speaker from Dhaka, Bangladesh. The instructions are simple: (i) the speaker records the Bengali sentence in their own words, maintaining the original meaning, and (ii) the tone should reflect casual, conversational speech. We then use an automatic speech recognition (ASR) tool\footnote{\url{https://cloud.google.com/speech-to-text}} to transcribe the spoken utterances to Bengali text, capturing both dialectal nuances and accent bias.

The dataset expansion process is shown in \cref{fig:workflow}, with the number of dialects per language cluster and the data transformation methods detailed in \cref{tab:lang}.

\subsection{LLM-as-a-Judge Consistency Evaluation}
Once we have the Toxigen human-annotated and expanded language variety dataset at hand, we move forward to the evaluation phase. Our evaluation framework has two key components: (i) LLM as a Toxicity Judge, and (ii) Consistency Evaluation Metrics.

\paragraph{LLM as a Toxicity Judge} We prompt instruction-tuned LLMs to assess the toxicity of statements in various dialects. As shown in Figure~\ref{fig:toxicity_instructions}, the prompt includes specific rubrics that help evaluators judge the severity of harmful language, target group identification, intent, and the impact of dialectal sensitivity on LLM responses. This structured approach ensures consistency across evaluations and captures nuanced insights into how LLMs respond to both standard and dialectal variations.
\begin{figure*}[ht]
\centering
\begin{tcolorbox}[colframe=blue!70, colback=gray!10, title=\textbf{LLM Toxicity Evaluation: Rubric and Instructions}, width=\textwidth]
\scriptsize 
\textbf{Instructions:} "You will be provided with a series of statements in various dialects. Your task is to evaluate the toxicity of each statement using the rubric below. Provide your judgment for each criterion (severity, target identification, and dialectal sensitivity). Assign a score between 1 (non-toxic) to 5 (severely toxic) for each statement. Format the output in JSON as shown below."

\textbf{Output Format:}
\begin{verbatim}
{
  "Input Statement": "{Input_statement}",
  "Severity": S1-S5,
  "Target Group Identification": T0-T3,
  "Dialectal Sensitivity Impact": D0-D3
}
\end{verbatim}

\vspace{0.5em}
\hrule
\vspace{0.5em}

\textbf{Evaluation Rubric:}
\begin{itemize}
    \item \textbf{Toxicity (S1-S5)}: Assign a score from 1 to 5 based on the level of harmfulness:
    \begin{itemize}
        \item \textbf{S1}: Not harmful (neutral or factual statement)
        \item \textbf{S2}: Slightly offensive (minor slurs or casual insults)
        \item \textbf{S3}: Moderately harmful (disrespectful or demeaning language)
        \item \textbf{S4}: Highly toxic (explicit hate speech or strong language)
        \item \textbf{S5}: Extremely toxic (incites violence, threats, or severe hate speech)
    \end{itemize}

    \item \textbf{Target Group Identification (T0-T3)}: Identify the explicit target and categorize the group affected:
    \begin{itemize}
        \item \textbf{T0}: No clear target (general or ambiguous)
        \item \textbf{T1}: Individual target (specific person or name)
        \item \textbf{T2}: Group target (demographic, gender, ethnicity)
        \item \textbf{T3}: Intersectional target (multiple demographics, e.g., race + gender)
    \end{itemize}

    \item \textbf{Dialectal Sensitivity Impact (D0-D3)}: Determine if the LLM’s perception changes due to dialectal wording:
    \begin{itemize}
        \item \textbf{D0}: No impact (judgment is consistent across dialects)
        \item \textbf{D1}: Minor impact (small variation, but no change in overall toxicity)
        \item \textbf{D2}: Noticeable impact (judgment changes moderately with dialectal wording)
        \item \textbf{D3}: Significant impact (substantial change in toxicity due to dialectal form)
    \end{itemize}
\end{itemize}

\textbf{Input Statement:} \texttt{\{Input\_statement\}}
\end{tcolorbox}
\caption{Rubric and instruction prompt for LLM-based toxicity evaluation across dialects.}
\label{fig:toxicity_instructions}
\end{figure*}

\paragraph{Consistency Evaluation Metrics} We argue that, a comprehensive multilingual LLM-as-a-judge evaluation must quantify three key dimensions of consistency: LLM-Human agreement, multilingual performance stability, and dialectal robustness. This is crucial as consistency across languages and dialects helps avoid bias toward specific linguistic groups, while ensuring fairness. To achieve this, we define the following consistency metrics:

\begin{itemize}[leftmargin=*,itemsep=0pt]
 \item \textbf{LLM-Human Consistency (\gapii{\text{lh}})}: This metric evaluates the alignment between the predictions made by the LLM and the human annotations across all language varieties, including the standard language. Essentially, it provides an aggregate measure of how consistently the LLM reproduces human judgments across different dialects and variations. The consistency is calculated by taking the average accuracy between the LLM predictions and human annotations for each of the $n$ language varieties:\\[-1.5em]
    \begin{align*}
    \gapii{\text{lh}} = \frac{\sum_{i=1}^{n} \text{Acc}(\text{LLM}_{i}, \text{Human})}{n}
    \end{align*}

    \item \textbf{Multilingual Consistency (\gapii{\text{ml}})}: This captures how consistently the LLM performs across multiple standard language varieties, measured as the variance of accuracy scores:\\[-1.5em]
    \begin{align*}
    \gapii{\text{ml}} = 1 - \text{Variance}(\text{Acc}_{\text{standard}})\\[-2em]
    \end{align*}
    A smaller variance indicates higher multilingual consistency.

    \item \textbf{Dialectal Consistency (\gapii{\text{dl}})}: This evaluates the accuracy difference between a standard variety (e.g., Standard English) and its corresponding dialects, assessing the model's robustness to shifts in dialectal variations:\\[-1.5em]
    \begin{align*}
   \gapii{\text{dl}} = 1 - \frac{1}{m} \sum_{i=1}^{m} \left| \text{Acc}_{\text{standard}} 
    - \text{Acc}_{\text{dialect}} \right| \\[-2em]
    \end{align*}
    where $m$ is the number of dialects per language. Smaller differences indicate better dialectal consistency.
\end{itemize}

\section{Experimental Setup}


We evaluate the performance of three large language models (LLMs)—\texttt{AYA-23-12B}~\cite{aryabumi2024aya}, \texttt{Phi-3 Mini-3.8B}~\cite{abdin2024phi3technicalreporthighly}, and \texttt{Mistral-NEMO-8B}~\cite{mistral_nemo}—to assess their capability in detecting toxicity across various language varieties. We compute accuracy and F1 scores, followed by evaluating consistency scores for all models and dialects to understand their performance and stability.

\begin{itemize}[leftmargin=*,itemsep=0pt]
    \item \textbf{\texttt{AYA-23-12B}}: A multilingual language model that focuses on 23 languages for enhanced depth, aiming to provide a balance between language diversity and modeling accuracy. AYA-23 builds on the previous Aya models, achieving superior performance for the covered languages compared to it's extended multilingual counterpart AYA-101~\cite{ustun-etal-2024-aya}.
    
    \item \textbf{\texttt{Phi-3 Mini-3.8B}}: A compact yet powerful language model, trained on 3.3 trillion tokens. It features multilingual long-context capabilities and is small enough to be deployed on mobile devices, making it a versatile choice for robust and safe language understanding.

   \item \textbf{\texttt{Mistral-NEMO-8B}}: This is a pruned 8B parameter version derived from the larger Mistral NeMo-12B model which was originally trained with quantization-aware techniques, enabling efficient FP8 inference without performance degradation. Mistral-NEMO-8B underwent advanced two-stage instruction fine-tuning and preference optimization.
\end{itemize}

For the remainder of this paper and the results section, we refer to \texttt{Mistral-NEMO-8B} as \texttt{NeMo}, \texttt{AYA-23-12B} as \texttt{Aya-23}, and \texttt{Phi-3 Mini-3.8B} as \texttt{Phi-3}.

\section{Results and Analysis}
\begin{figure*}
    \centering
    \includegraphics[width=\textwidth]{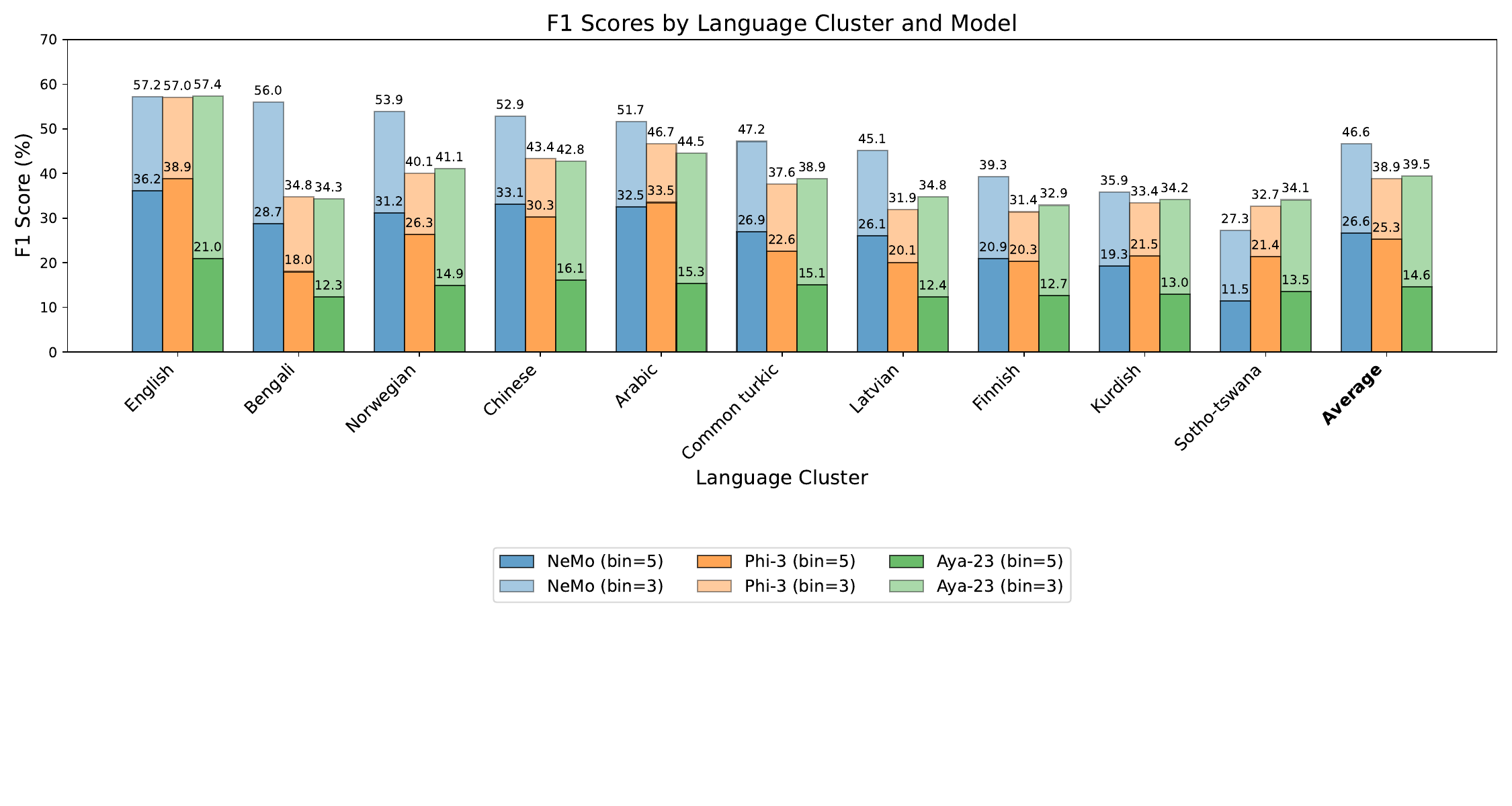}
    \vspace{-6em}
    \caption{We compute F1 scores for each language cluster by averaging over all dialects. The original ToxiGen intent labels, which range continuously from 1 to 5, are converted into bins of 3 and 5 for evaluation. The results indicate an overall low agreement (below 50\%) with human annotations, where \nemo{} has the highest scores. The overall performance tends to decrease for low-resource language varieties}
    \label{fig:summary_models}
\end{figure*}


\paragraph{Broad model comparisons} In \cref{fig:summary_models}, we report the average F1 scores for each language group and for all three models. For more interpretable evaluation, we converted the 10 continuous point Toxigen intent labels and the LLM-predicted toxicity into bins of 3 and 5, then computed accuracy and F1 scores in a multiclass format. For aggregated F1 scores (avg.) across language clusters, \nemo{} consistently achieves higher performance compared to Phi-3 and Aya, especially in the 3-class evaluation setup. \nemo’s superior performance can be attributed to its exposure to multilingual instruction tuning, which helps it handle diverse languages and dialects more effectively. However, while \ayaa{} is also tuned on 23 languages, it still struggles to outperform \nemo{} and, at times, even lags behind the much smaller \phii{} model. Despite their strengths, the overall agreement between human labels and all three models remains low, indicating a significant gap in aligning model predictions with human judgment.

\paragraph{Results across language clusters} The models exhibited varying performance across different language clusters. In high-resource languages like English, Norwegian, and Chinese, \nemo{} consistently performed well, with \ayaa{} also showing strong results. However, in lower-resource languages such as Sotho-Tswana and Kurdish, all models struggled, highlighting the performance gap in these underrepresented languages. Interestingly, in some of these lower-resource languages, \phii, despite its smaller size, performed comparably to \nemo{}, narrowing the performance disparity in certain clusters.
\begin{table*}[t]
\small
\centering
\begin{tabular}{p{3.5cm}p{3.5cm}p{3.5cm}p{3.5cm}}
\toprule
Language Cluster & \nemo{} & \ayaa{} & \phii{} \\
\midrule
Arabic & \underline{Standard Arabic} & North Mesopotamian Arabic & North Mesopotamian Arabic \\
Bengali & \underline{Standard} & \underline{Standard} & Dhaka \\
Chinese & Classical-Middle-Modern Sinitic (Simplified) & Classical-Middle-Modern Sinitic (Simplified) & Classical-Middle-Modern Sinitic (Simplified) \\
Common Turkic & \underline{Central Oghuz} & \underline{Central Oghuz} & North Azerbaijani \\
English & \underline{Standard} & \underline{Standard} & \underline{Standard} \\
Finnish & \underline{Finnish} & Lansi-Uusimaa & Keski-Karjala \\
Kurdish & Northern Kurdish & \underline{Central Kurdish} & Northern Kurdish \\
Latvian & \underline{Latvian} & \underline{Latvian} & East Latvian \\
Norwegian & Norwegian Nynorsk & Norwegian Nynorsk & \underline{Norwegian Bokmal} \\
Sotho-Tswana & Southern Sotho & Southern Sotho & Southern Sotho \\
\midrule
Cluster Representative (\%) & 60.00\% & 50.00\% & 20.00\% \\
\bottomrule
\end{tabular}
\caption{Comparison of Highest Scoring Varieties for Each Language Cluster. The highest performing variety for each cluster is not always the cluster representative, which generally represents the high-resource or standard variety. Underlined dialects indicate the cluster representative.}
\label{tab:highest_scoring_dialects}
\end{table*}

\paragraph{Highest Scoring Varieties and Dialectal Gap} In \cref{tab:highest_scoring_dialects}, we present the highest scoring dialects for each language cluster. The results show a gap between the best performing varieties and the standard or high-resource dialects, which are often considered representative of their clusters. The highest scoring variety for each model does not always align with the standard dialect (underlined). This indicates that models may perform better on less common or non-standard varieties. These findings highlight a potential dialectal gap, where high-resource dialects are not necessarily the most favorable in terms of model performance.

\begin{table}[t]
\small
\centering
\begin{tabular}{lrrr}
\toprule
Model & LLM-Human & Multilingual & Dialectal  \\
\midrule
\phii{} & 49.83 & 99.60 & 98.40 \\
\ayaa{} & 49.14 & 99.81 & 98.50 \\
\nemo{} & 48.48 & 98.75 & 91.75 \\
\midrule
Average & 49.15 & 99.39 & 96.22 \\
\bottomrule
\end{tabular}
\vspace{-.5em}
\caption{Model level Consistency Scores (\%, Global)}
\label{tab:consistency_scores}
\vspace{-1em}
\end{table}

\begin{table}[t]
\small
\centering
\begin{tabular}{lrrr}
\toprule
Model & \nemo{} & \phii{} & \ayaa{} \\
Language Cluster &  &  &  \\
\midrule
Arabic & 95.7 & 98.0 & 98.7 \\
Bengali & 93.6 & 97.7 & 99.2 \\
Chinese & 97.1 & 99.5 & 97.7 \\
Common Turkic & 91.7 & 97.5 & 94.9 \\
English & 96.1 & 97.8 & 97.7 \\
Finnish & 87.5 & 98.9 & 99.0 \\
Latvian & 86.0 & 99.9 & 98.3 \\
Kurdish & 95.7 & 97.0 & 99.7 \\
Norwegian & 99.6 & 96.5 & 99.5 \\
Sotho-Tswana & 97.4 & 98.7 & 99.9 \\
\midrule
Average & 94.0 & 98.2 & 98.5 \\
\bottomrule
\end{tabular}
\vspace{-.5em}
\caption{Dialectal Consistency Scores for Different Language Clusters (\%). \phii{} and \ayaa{} show superior and consistent dialectal consistency across all language clusters compared to \nemo, which struggles particularly with Finnish and Latvian.}
\label{tab:dialectal_consistency_scores}
\vspace{-1em}
\end{table}

\paragraph{LLM Consistency Evaluation} As shown in \cref{tab:consistency_scores}, the three global consistency scores for all models reveal that dialectal consistency is slightly lower than multilingual consistency, though the gap between the two is relatively small. This indicates that LLMs can handle both multilingual and dialectal variations reasonably well. However, \nemo{} notably struggles more with dialectal robustness compared to the other models. However, the primary challenge remains improving LLM-human agreement, which consistently shows the weakest performance across all models. 

We also report the individual dialectal consistency within each language cluster in \cref{tab:dialectal_consistency_scores}. We observe, \nemo{} shows drops in dialectal consistency, particularly with Finnish (87.5\%) and Latvian (86.0\%), both high-resource language groups. It also struggles with Bengali and Common Turkic, indicating recurring challenges in managing dialectal variations within these clusters. The structural and linguistic diversity of these languages suggests that NeMo may be sensitive to complex dialectal features regardless of resource level, highlighting potential limitations in handling diverse morphological and syntactic patterns across dialects. In contrast, \phii{} and \ayaa{} exhibit stronger and more balanced performance across all language clusters, with consistently high dialectal consistency scores (averages of 98.2\% and 98.5\%, respectively). This suggests that both models are better equipped to handle variations within dialects, regardless of the resource level or structural characteristics of the language cluster.


\begin{figure*}[!htpb]
    \centering
    \includegraphics[width=\textwidth]{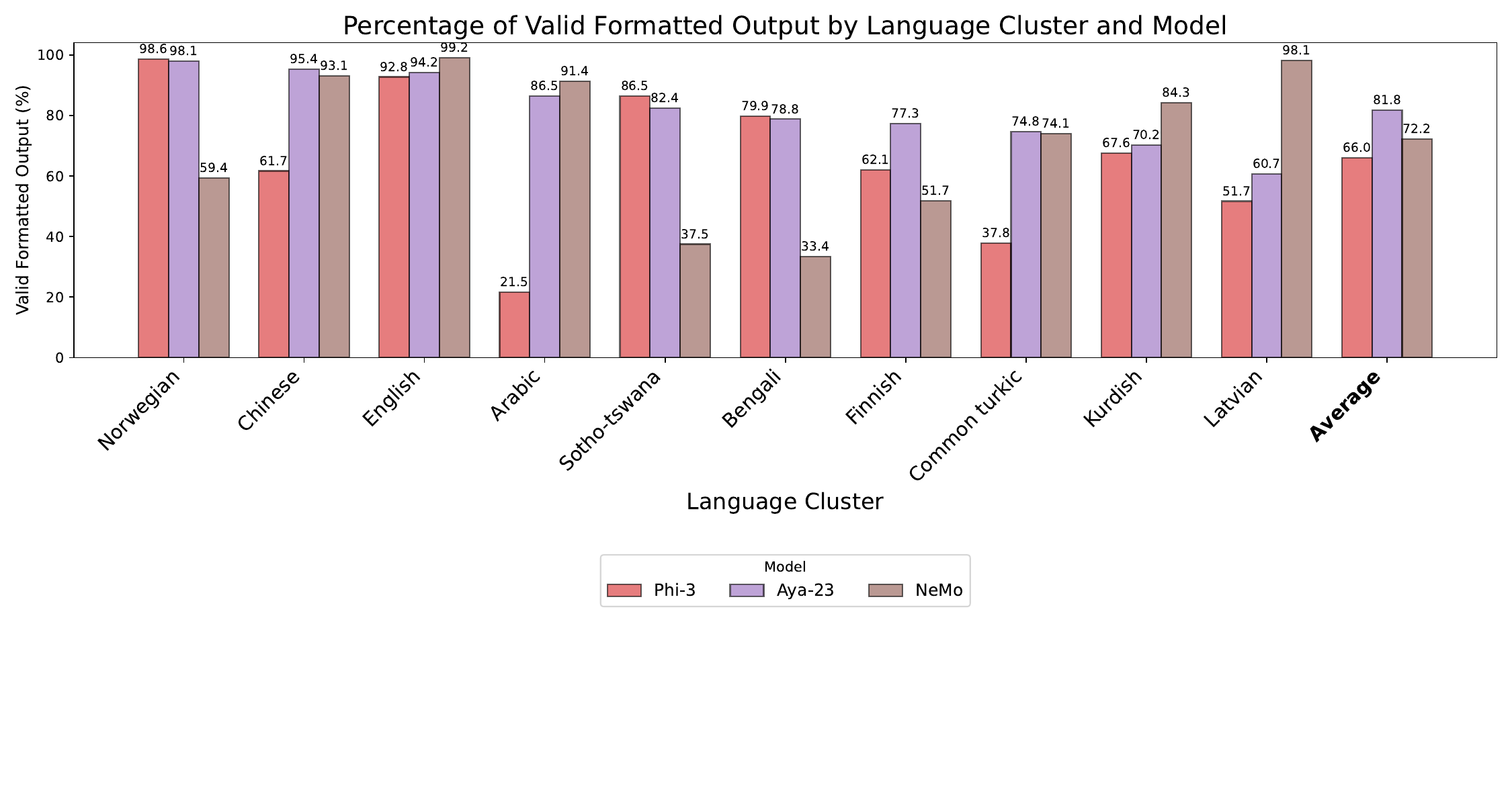}
    \vspace{-6em}
    \caption{This bar plot shows how different language models (\phii, \ayaa, and \nemo) perform in terms of valid formatted output across multiple language clusters when given multilingual instructions. While all models achieve high performance for English, their performance varies significantly for other languages. Aya-23 generally performs better, whereas Phi-3 struggles more across most languages.}
    \label{fig:valid_format}
\end{figure*}

\paragraph{LLMs Struggle with Multilingual Instruction Following:} Both \ayaa{} and \nemo{} show strong multilingual instruction-following abilities in most cases, as indicated by the percentage of valid JSON-formatted outputs in \cref{fig:valid_format}. \ayaa{} consistently performs well across most language clusters, which can be attributed to its multilingual instruction tuning on 23 languages and its higher parameter count compared to the other models. All three models achieve over 92\% success in instruction following and output formatting in English, but their performance declines when dealing with other languages. Interestingly, the extent of degradation varies across language clusters for each model. For example, \nemo{} also performs well and in some cases surpasses Aya but its performance drops significantly in Bengali (a drop of 46.5 percentage points from the highest performing model) and Sotho (a drop of 49.0 percentage points from the highest performing model). This could further explain NeMo's higher Bengali score in \cref{fig:summary_models}, as the lower number of example count skewed the evaluation, making NeMo's results in Bengali appear better (56\%) wheres the two other models struggled to get decent scores (34.8\% and 34.3\%).

\begin{table*}[]
    \centering
    \small
\begin{tabular}{p{4cm}p{4cm}p{4cm}l}
\toprule
Language Cluster & Dialect (Min Impact) & Dialect (Max Impact) & $\Delta$ (Max - Min) \\
\midrule
Arabic & \underline{Standard arabic} (38.95) & Moroccan Arabic (56.41) & 17.46 \\
Bengali & \underline{Standard} (60.45) & Dhaka (65.02) & 4.58 \\
Chinese & Classical-Middle-Modern Sinitic (Traditional) (29.10) & \underline{Cantonese} (54.63) & 25.53 \\
Common Turkic & \underline{Central oghuz} (34.03) & South Azerbaijani (75.30) & 41.27 \\
English & \underline{Standard} (22.79) & Southeast American Enclave (61.81) & 39.02 \\
Finnish & \underline{Finnish} (35.70) & PohjoinenVarsinais-Suomi (65.55) & 29.86 \\
High German & \underline{Latvian} (45.21) & East Latvian (66.00) & 20.79 \\
Kurdish & Northern Kurdish (67.65) & \underline{Central kurdish} (76.71) & 9.05 \\
Norwegian & Norwegian Nynorsk (42.69) & \underline{Norwegian bokmal} (45.53) & 2.84 \\
Sotho-Tswana & Southern Sotho (76.30) & \underline{Northern sotho} (79.76) & 3.46 \\
\midrule
Cluster Representatives (\%) & 60.00\% & 40.00\% &  \\
\bottomrule
\end{tabular}
    \caption{Comparison of varieties with the minimum and maximum dialectal impact acknowledgment by LLMs while judging each language cluster. The percentage points are averaged across the three LLMs. Cluster representatives are underlined, indicating minimal dialectal sensitivity. The $\Delta$ represents the difference between the maximum and minimum impact percentages. The key takeaway is that standard varieties show minimal impact, non-standard varieties show maximum impact, and ambiguity arises when defining a single standard variety is not straightforward (e.g. Northern and Southern Sotho).}

    \label{tab:max_min_sensitivity}
\end{table*}

\paragraph{LLM Perception of Dialectal Sensitivity:} We performed an ablation task to gain a deeper understanding of how LLMs, acting as judges, perceive dialectal sensitivity when assessing toxicity. The goal of this task was to determine whether, according to LLMs, dialectal variations affect their judgment of toxicity. The model either acknowledges or denies the impact, and the results are summarized in the table. Ideally, the LLMs should recognize that standard language varieties contain minimal dialectal cues compared to their dialectal counterpart.

Here’s the specific part of the prompt provided to LLMs (see \cref{fig:toxicity_instructions} for the detailed prompt):

\begin{quote} \item \textbf{Dialectal Sensitivity Impact (D0-D3)}: Determine if the LLM’s perception changes due to dialectal wording given the input statement: \begin{itemize}[leftmargin=*,itemsep=0pt] \item \textbf{D0}: No impact (judgment is consistent across dialects) \item \textbf{D1}: Minor impact (small variation, but no change in overall toxicity) \item \textbf{D2}: Noticeable impact (judgment changes moderately with dialectal wording) \item \textbf{D3}: Significant impact (substantial change in toxicity due to dialectal form) \end{itemize} \textbf{Input Statement:} \texttt{\{Input\_statement\}} \end{quote}

In \cref{{tab:max_min_sensitivity}}, we report the language variety with the highest and lowest percentage of acknowledgment (i.e., the percentage of instances where the model recognized any of the D1, D2, or D3 levels of impact) for each language cluster. Ideally, when serving as a judge, the LLM should recognize that standard varieties, such as Standard English, contain no significant dialectal nuances and should respond with D0 (no impact) or show a lower percentage of acknowledgment for dialectal impact (Minimal Impact). \cref{{tab:max_min_sensitivity}} shows that this is generally the case for standard varieties.

We observed that the percentage of LLMs acknowledging an impact was significantly lower when the text was presented in standard varieties, such as Standard English, Finnish, and Standard Bengali, with percentages averaging around 60\%. This suggests that, according to the LLMs, their predictions as a judge show minimal sensitivity to dialectal differences in these cases. On the other hand, LLMs acknowledge greater sensitivity when judging non-standard dialects, indicating a stronger recognition of dialectal nuances in these variations.

Even when maximum impact occurs in the case of a cluster representative, such as Northern Sotho, Norwegian Bokmal, and Central Kurdish, these cases involve closely related languages where defining a single standard form is not ideal. In such instances, the delta difference between the maximum and minimum impacts remains below 10\%. This suggests that while LLMs can effectively detect dialectal differences, their sensitivity is influenced by the lack of a clearly defined standard dialect. Therefore, the more ambiguous the standard form within a cluster, the lower the impact of dialectal variation on the model's judgment.

\section{Conclusion and Future Work}

We propose a holistic LLM robustness evaluation framework for handling toxicity across language varieties. Our findings suggest, a notable gap remains between model predictions and human judgment, emphasizing the need for improvements in alignment. Additionally, LLMs tend to be more sensitive to non-standard dialects, indicating that further advancements are required to enhance their consistency across diverse language varieties. We aim to further expand our dataset by incorporating more utterance-based dialects and introducing new perturbation methods, leveraging LLMs' understanding of dialectal variations.


\section*{Limitations}
At this point, this study mostly contains synthetic and machine-translated dialectal varieties except for one set of spoken utterances (Bengali-Dhaka).
While it would be ideal to conduct this study on authentic data, such data are not easily available and they are expensive to collect. This low percentage of real-world dialectal examples is a limitation we hope to address in the future.



\section*{Acknowledgments}
This material is based upon work supported by the US National Science Foundation under Grants No. IIS-2125466 and IIS-2327143.

\bibliography{anthology,custom}
\bibliographystyle{acl_natbib}

\appendix
\onecolumn

\newpage
\section*{Appendix}
\label{app:start}


\subsection*{Detailed Evaluation Results}
\label{app:result_tables}
\begin{table}
\small
\begin{tabular}{llrrrr}
\toprule
 &  & ACC(bin=3) & ACC(bin=5) & F1(bin=3) & F1(bin=5) \\
Language Cluster & Dialect &  &  &  &  \\
\midrule
\multirow[t]{9}{*}{Arabic} & North mesopotamian arabic & 53.10 & 24.70 & 46.70 & 15.30 \\
 & Ta'izzi-adeni arabic & 51.30 & 24.70 & 44.60 & 15.20 \\
 & Tunisian arabic & 50.70 & 24.20 & 43.40 & 14.70 \\
 & South levantine arabic & 52.90 & 25.70 & 46.70 & 16.50 \\
 & Levantine arabic (a:north) & 48.60 & 23.40 & 41.70 & 14.40 \\
 & Standard arabic & 51.70 & 24.40 & 45.20 & 15.40 \\
 & Najdi arabic & 52.00 & 24.20 & 45.70 & 15.00 \\
 & Moroccan arabic & 49.40 & 24.90 & 41.90 & 15.10 \\
 & Egyptian arabic & 51.10 & 24.80 & 44.80 & 15.80 \\
\cmidrule{1-6}
\multirow[t]{3}{*}{Chinese} & Cantonese & 50.30 & 25.20 & 43.90 & 15.70 \\
 & Classical-middle-modern sinitic (simplified) & 52.00 & 26.40 & 45.50 & 17.30 \\
 & Classical-middle-modern sinitic (traditional) & 47.40 & 23.90 & 38.90 & 15.40 \\
\cmidrule{1-6}
\multirow[t]{24}{*}{Finnish} & Finnish & 45.60 & 24.20 & 33.40 & 12.70 \\
 & Pohjois-satakunta & 44.70 & 24.80 & 33.10 & 13.30 \\
 & Keski-karjala & 45.60 & 24.00 & 33.40 & 12.10 \\
 & Kainuu & 44.00 & 25.20 & 32.30 & 14.10 \\
 & Etela-pohjanmaa & 44.10 & 23.70 & 33.00 & 12.80 \\
 & Etela-satakunta & 44.40 & 24.30 & 33.00 & 13.10 \\
 & Pohjois-savo & 45.50 & 24.10 & 33.30 & 12.40 \\
 & Pohjois-karjala & 44.00 & 23.40 & 32.40 & 12.40 \\
 & Keski-pohjanmaa & 44.20 & 24.50 & 32.60 & 13.20 \\
 & Kaakkois-hame & 45.60 & 24.80 & 34.20 & 13.20 \\
 & Pohjoinenkeski-suomi & 45.50 & 23.90 & 34.20 & 12.70 \\
 & Pohjois-pohjanmaa & 43.40 & 24.40 & 32.30 & 13.30 \\
 & Pohjoinenvarsinais-suomi & 44.80 & 24.60 & 32.90 & 12.80 \\
 & Etela-karjala & 44.70 & 24.90 & 32.30 & 13.20 \\
 & Lansi-uusimaa & 45.60 & 25.40 & 34.00 & 13.80 \\
 & Inkerinsuomalaismurteet & 44.70 & 23.70 & 32.10 & 11.80 \\
 & Lantinenkeski-suomi & 42.90 & 23.30 & 31.20 & 12.20 \\
 & Lansi-satakunta & 45.10 & 23.60 & 33.90 & 12.70 \\
 & Etela-savo & 44.00 & 23.30 & 31.70 & 11.90 \\
 & Lansipohja & 43.50 & 23.50 & 32.00 & 12.50 \\
 & Pohjois-hame & 45.40 & 23.70 & 33.80 & 12.10 \\
 & Etelainenkeski-suomi & 44.40 & 24.80 & 32.60 & 12.90 \\
 & Etela-hame & 43.90 & 23.90 & 31.70 & 12.20 \\
 & Perapohjola & 45.20 & 23.70 & 34.50 & 12.50 \\
\cmidrule{1-6}
\multirow[t]{2}{*}{Kurdish} & Central kurdish & 46.10 & 23.40 & 32.70 & 11.80 \\
 & Northern kurdish & 45.90 & 24.70 & 35.80 & 14.10 \\
\cmidrule{1-6}
\multirow[t]{2}{*}{Norwegian} & Norwegian nynorsk & 49.80 & 24.20 & 40.70 & 14.20 \\
 & Norwegian bokmal & 49.20 & 24.80 & 41.60 & 15.50 \\
\cmidrule{1-6}
\multirow[t]{2}{*}{High\_german} & East latvian & 45.90 & 24.60 & 34.40 & 13.00 \\
 & Latvian & 47.60 & 24.30 & 35.30 & 11.80 \\
\cmidrule{1-6}
\multirow[t]{11}{*}{English} & Standard & 61.20 & 30.80 & 59.90 & 24.20 \\
 & Southeast american enclave & 58.10 & 29.00 & 56.10 & 20.80 \\
 & Chicano & 60.80 & 29.70 & 59.50 & 22.30 \\
 & Nigerian & 59.80 & 29.20 & 58.40 & 21.90 \\
 & African american vernacular & 57.10 & 27.30 & 55.30 & 20.20 \\
 & Appalachian & 58.80 & 28.90 & 57.20 & 21.40 \\
 & Australian & 60.80 & 28.60 & 59.40 & 20.70 \\
 & Colloquial singapore & 59.10 & 28.30 & 57.10 & 20.90 \\
 & Hong kong & 57.60 & 27.10 & 55.20 & 19.40 \\
 & Indian & 57.90 & 26.30 & 56.00 & 18.80 \\
 & Irish & 59.00 & 28.30 & 57.20 & 20.40 \\
\cmidrule{1-6}
\multirow[t]{2}{*}{Sotho-tswana} & Northern sotho & 47.60 & 25.70 & 34.90 & 13.50 \\
 & Southern sotho & 47.70 & 27.10 & 33.40 & 13.40 \\
\cmidrule{1-6}
\multirow[t]{2}{*}{Bengali} & Dhaka & 45.50 & 22.90 & 33.50 & 11.90 \\
 & Standard & 46.30 & 22.50 & 35.00 & 12.80 \\
\cmidrule{1-6}
\multirow[t]{3}{*}{Common\_turkic} & Central oghuz & 51.70 & 25.60 & 44.90 & 17.10 \\
 & South azerbaijani & 46.90 & 25.30 & 35.00 & 13.60 \\
 & North azerbaijani & 46.30 & 24.20 & 36.70 & 14.60 \\
\cmidrule{1-6}
\multirow[t]{2}{*}{Average} & - & 49.10 & 25.20 & 40.40 & 15.00 \\
\cmidrule{1-6}
\bottomrule
\end{tabular}
\caption{Evaluation Results for aya-23-8B}
\label{tab:aya-23-8B}
\end{table}

\begin{table}
\centering
\small
\begin{tabular}{llrrrr}
\toprule
 &  & ACC(bin=3) & ACC(bin=5) & F1(bin=3) & F1(bin=5) \\
Language Cluster & Dialect &  &  &  &  \\
\midrule
\multirow[t]{9}{*}{Arabic} & North mesopotamian arabic & 52.50 & 36.80 & 49.80 & 32.60 \\
 & Ta'izzi-adeni arabic & 55.30 & 37.70 & 53.10 & 33.50 \\
 & Tunisian arabic & 52.60 & 33.80 & 50.80 & 30.60 \\
 & South levantine arabic & 53.30 & 35.50 & 51.30 & 32.20 \\
 & Levantine arabic (a:north) & 53.20 & 34.60 & 51.00 & 32.00 \\
 & Standard arabic & 57.80 & 39.10 & 55.10 & 34.10 \\
 & Najdi arabic & 55.20 & 36.90 & 53.40 & 32.60 \\
 & Moroccan arabic & 51.40 & 35.00 & 48.80 & 32.10 \\
 & Egyptian arabic & 54.50 & 35.60 & 52.30 & 32.40 \\
\cmidrule{1-6}
\multirow[t]{3}{*}{Chinese} & Cantonese & 57.40 & 38.90 & 54.70 & 33.80 \\
 & Classical-middle-modern sinitic (simplified) & 58.50 & 41.00 & 55.10 & 35.20 \\
 & Classical-middle-modern sinitic (traditional) & 52.70 & 36.70 & 48.70 & 30.20 \\
\cmidrule{1-6}
\multirow[t]{24}{*}{Finnish} & Finnish & 53.70 & 36.20 & 51.40 & 31.00 \\
 & Pohjois-satakunta & 42.90 & 25.40 & 41.60 & 23.40 \\
 & Keski-karjala & 43.50 & 26.20 & 42.50 & 24.70 \\
 & Kainuu & 42.60 & 24.30 & 41.00 & 22.20 \\
 & Etela-pohjanmaa & 42.80 & 27.00 & 41.60 & 25.20 \\
 & Etela-satakunta & 43.70 & 23.70 & 42.70 & 22.20 \\
 & Pohjois-savo & 43.90 & 24.30 & 43.20 & 22.10 \\
 & Pohjois-karjala & 42.40 & 24.60 & 41.10 & 22.10 \\
 & Keski-pohjanmaa & 41.40 & 23.50 & 40.10 & 21.20 \\
 & Kaakkois-hame & 39.50 & 21.30 & 37.40 & 18.00 \\
 & Pohjoinenkeski-suomi & 40.80 & 20.10 & 37.40 & 15.60 \\
 & Pohjois-pohjanmaa & 40.80 & 23.70 & 38.10 & 20.60 \\
 & Pohjoinenvarsinais-suomi & 39.10 & 21.50 & 36.20 & 19.00 \\
 & Etela-karjala & 39.30 & 25.50 & 36.90 & 22.20 \\
 & Lansi-uusimaa & 39.10 & 21.40 & 35.70 & 16.80 \\
 & Inkerinsuomalaismurteet & 36.90 & 23.10 & 34.20 & 19.20 \\
 & Lantinenkeski-suomi & 41.60 & 24.20 & 38.70 & 19.70 \\
 & Lansi-satakunta & 37.80 & 18.70 & 34.00 & 14.60 \\
 & Etela-savo & 42.00 & 24.00 & 39.60 & 19.30 \\
 & Lansipohja & 41.50 & 23.80 & 39.00 & 21.50 \\
 & Pohjois-hame & 40.60 & 23.90 & 37.10 & 21.20 \\
 & Etelainenkeski-suomi & 40.80 & 20.80 & 36.80 & 16.30 \\
 & Etela-hame & 42.10 & 25.10 & 38.20 & 20.40 \\
 & Perapohjola & 41.50 & 26.00 & 39.10 & 22.40 \\
\cmidrule{1-6}
\multirow[t]{2}{*}{Kurdish} & Central kurdish & 34.00 & 20.60 & 32.80 & 18.00 \\
 & Northern kurdish & 38.30 & 22.50 & 39.00 & 20.60 \\
\cmidrule{1-6}
\multirow[t]{2}{*}{Norwegian} & Norwegian nynorsk & 56.80 & 38.50 & 54.50 & 33.50 \\
 & Norwegian bokmal & 56.40 & 33.50 & 53.30 & 28.90 \\
\cmidrule{1-6}
\multirow[t]{2}{*}{High\_german} & East latvian & 39.00 & 22.60 & 38.90 & 20.20 \\
 & Latvian & 53.00 & 35.40 & 51.30 & 32.00 \\
\cmidrule{1-6}
\multirow[t]{11}{*}{English} & Standard & 63.50 & 44.40 & 60.10 & 38.70 \\
 & Southeast american enclave & 58.50 & 38.90 & 55.60 & 35.00 \\
 & Chicano & 61.60 & 41.80 & 58.70 & 36.10 \\
 & Nigerian & 60.30 & 40.90 & 57.60 & 36.50 \\
 & African american vernacular & 59.40 & 39.40 & 56.50 & 35.70 \\
 & Appalachian & 60.00 & 41.00 & 57.20 & 35.60 \\
 & Australian & 58.70 & 40.70 & 55.90 & 36.30 \\
 & Colloquial singapore & 58.30 & 39.40 & 56.00 & 36.30 \\
 & Hong kong & 59.20 & 39.40 & 57.10 & 35.70 \\
 & Indian & 58.90 & 38.70 & 56.40 & 34.40 \\
 & Irish & 60.70 & 41.10 & 57.90 & 37.90 \\
\cmidrule{1-6}
\multirow[t]{2}{*}{Sotho-tswana} & Northern sotho & 27.10 & 13.70 & 25.40 & 9.60 \\
 & Southern sotho & 29.70 & 16.10 & 29.20 & 13.40 \\
\cmidrule{1-6}
\multirow[t]{2}{*}{Bengali} & Dhaka & 54.50 & 28.10 & 52.00 & 23.50 \\
 & Standard & 60.90 & 37.60 & 60.00 & 33.90 \\
\cmidrule{1-6}
\multirow[t]{3}{*}{Common\_turkic} & Central oghuz & 53.90 & 37.40 & 52.60 & 32.20 \\
 & South azerbaijani & 38.70 & 21.50 & 37.40 & 18.60 \\
 & North azerbaijani & 52.50 & 34.20 & 51.50 & 30.00 \\
\cmidrule{1-6}
\multirow[t]{2}{*}{Average} & - & 48.50 & 30.50 & 46.20 & 26.70 \\
\cmidrule{1-6}
\bottomrule
\end{tabular}
\caption{Evaluation Results for Mistral-NeMo-Minitron-8B-Instruct}
\label{tab:Mistral-NeMo-Minitron-8B-Instruct}
\end{table}

\begin{table}
\centering
\small
\begin{tabular}{llrrrr}
\toprule
 &  & ACC(bin=3) & ACC(bin=5) & F1(bin=3) & F1(bin=5) \\
Language Cluster & Dialect &  &  &  &  \\
\midrule
\multirow[t]{9}{*}{Arabic} & North mesopotamian arabic & 56.50 & 39.60 & 50.80 & 36.60 \\
 & Ta'izzi-adeni arabic & 51.10 & 34.20 & 43.40 & 29.50 \\
 & Tunisian arabic & 53.00 & 35.90 & 45.50 & 31.50 \\
 & South levantine arabic & 52.70 & 35.30 & 45.70 & 31.40 \\
 & Levantine arabic (a:north) & 54.90 & 39.90 & 48.10 & 37.10 \\
 & Standard arabic & 53.10 & 34.90 & 46.80 & 31.30 \\
 & Najdi arabic & 48.90 & 39.40 & 41.70 & 34.10 \\
 & Moroccan arabic & 54.00 & 37.40 & 47.90 & 33.60 \\
 & Egyptian arabic & 56.20 & 39.90 & 50.60 & 36.80 \\
\cmidrule{1-6}
\multirow[t]{3}{*}{Chinese} & Cantonese & 50.10 & 30.50 & 45.10 & 28.10 \\
 & Classical-middle-modern sinitic (simplified) & 50.60 & 37.00 & 43.60 & 33.20 \\
 & Classical-middle-modern sinitic (traditional) & 49.50 & 34.80 & 41.50 & 29.50 \\
\cmidrule{1-6}
\multirow[t]{24}{*}{Finnish} & Finnish & 43.90 & 28.90 & 30.70 & 20.70 \\
 & Pohjois-satakunta & 44.90 & 28.20 & 31.90 & 20.50 \\
 & Keski-karjala & 46.60 & 29.80 & 32.40 & 21.60 \\
 & Kainuu & 45.00 & 27.40 & 31.50 & 20.00 \\
 & Etela-pohjanmaa & 44.90 & 28.30 & 30.20 & 19.70 \\
 & Etela-satakunta & 43.60 & 27.70 & 30.30 & 20.40 \\
 & Pohjois-savo & 46.40 & 30.40 & 33.40 & 22.60 \\
 & Pohjois-karjala & 46.30 & 26.60 & 34.20 & 20.00 \\
 & Keski-pohjanmaa & 43.50 & 26.30 & 29.90 & 18.60 \\
 & Kaakkois-hame & 45.60 & 28.60 & 31.50 & 20.60 \\
 & Pohjoinenkeski-suomi & 45.60 & 28.80 & 31.70 & 21.00 \\
 & Pohjois-pohjanmaa & 43.90 & 24.90 & 31.00 & 18.40 \\
 & Pohjoinenvarsinais-suomi & 44.00 & 26.80 & 30.90 & 19.30 \\
 & Etela-karjala & 44.50 & 28.80 & 30.40 & 20.70 \\
 & Lansi-uusimaa & 44.70 & 27.00 & 32.00 & 19.80 \\
 & Inkerinsuomalaismurteet & 46.20 & 27.60 & 32.60 & 20.00 \\
 & Lantinenkeski-suomi & 43.40 & 27.60 & 29.40 & 19.60 \\
 & Lansi-satakunta & 44.30 & 28.00 & 31.30 & 20.60 \\
 & Etela-savo & 44.80 & 28.50 & 30.80 & 20.50 \\
 & Lansipohja & 44.30 & 27.70 & 31.20 & 20.40 \\
 & Pohjois-hame & 44.70 & 29.90 & 31.40 & 21.80 \\
 & Etelainenkeski-suomi & 44.50 & 28.40 & 31.40 & 21.10 \\
 & Etela-hame & 45.70 & 27.80 & 32.40 & 20.10 \\
 & Perapohjola & 45.40 & 27.40 & 31.40 & 19.50 \\
\cmidrule{1-6}
\multirow[t]{2}{*}{Kurdish} & Central kurdish & 44.20 & 26.20 & 32.80 & 20.80 \\
 & Northern kurdish & 47.20 & 29.50 & 33.90 & 22.20 \\
\cmidrule{1-6}
\multirow[t]{2}{*}{Norwegian} & Norwegian nynorsk & 48.80 & 32.20 & 37.40 & 24.40 \\
 & Norwegian bokmal & 52.20 & 35.00 & 42.90 & 28.10 \\
\cmidrule{1-6}
\multirow[t]{2}{*}{High\_german} & East latvian & 45.20 & 27.20 & 31.80 & 19.70 \\
 & Latvian & 45.10 & 28.40 & 32.10 & 20.50 \\
\cmidrule{1-6}
\multirow[t]{11}{*}{English} & Standard & 63.90 & 45.60 & 58.80 & 43.10 \\
 & Southeast american enclave & 62.20 & 39.20 & 57.50 & 37.90 \\
 & Chicano & 63.30 & 43.20 & 58.50 & 41.50 \\
 & Nigerian & 62.00 & 40.40 & 57.20 & 39.10 \\
 & African american vernacular & 60.00 & 38.70 & 55.40 & 37.50 \\
 & Appalachian & 62.90 & 41.60 & 58.00 & 39.80 \\
 & Australian & 63.70 & 40.30 & 58.90 & 38.90 \\
 & Colloquial singapore & 60.70 & 40.30 & 56.00 & 39.00 \\
 & Hong kong & 58.30 & 38.10 & 53.20 & 36.50 \\
 & Indian & 60.40 & 38.60 & 55.60 & 36.90 \\
 & Irish & 63.40 & 39.50 & 58.20 & 38.00 \\
\cmidrule{1-6}
\multirow[t]{2}{*}{Sotho-tswana} & Northern sotho & 45.80 & 28.80 & 32.40 & 22.40 \\
 & Southern sotho & 47.10 & 27.90 & 33.00 & 20.40 \\
\cmidrule{1-6}
\multirow[t]{2}{*}{Bengali} & Dhaka & 43.70 & 21.00 & 36.60 & 16.80 \\
 & Standard & 41.40 & 23.80 & 33.10 & 19.20 \\
\cmidrule{1-6}
\multirow[t]{3}{*}{Common\_turkic} & Central oghuz & 45.40 & 27.90 & 35.30 & 21.80 \\
 & South azerbaijani & 47.70 & 28.10 & 39.10 & 23.80 \\
 & North azerbaijani & 48.10 & 27.80 & 38.30 & 22.20 \\
\cmidrule{1-6}
\multirow[t]{2}{*}{Average} & - & 49.80 & 32.00 & 39.80 & 26.50 \\
\cmidrule{1-6}
\bottomrule
\end{tabular}
\caption{Evaluation Results for Phi-3-mini-4k-instruct}
\label{tab:Phi-3-mini-4k-instruct}
\end{table}

This section presents the detailed result tables (\crefrange{tab:aya-23-8B}{tab:Phi-3-mini-4k-instruct}) summarizing the performance of each model across different languages and dialects. We report metrics such as accuracy, F1 scores (for both bin=3 and bin=5 classifications), and the consistency scores for multilingual, dialectal, and LLM-human agreement. 

\subsection*{Dialect Sensitivity Evaluation}
\label{app:dialect_sensitivity}
\begin{table}[]
    \centering
    \small
\begin{tabular}{llrrrr}
\toprule
 &  & Phi & Aya & NeMo & Average \\
Language & Dialect &  &  &  &  \\
\midrule
\multirow[t]{9}{*}{Arabic} & Standard Arabic & 34.29 & 37.44 & 45.13 & 38.95 \\
 & Najdi Arabic & 35.50 & 40.84 & 53.97 & 43.44 \\
 & North Mesopotamian Arabic & 33.33 & 44.71 & 57.38 & 45.14 \\
 & Ta'izzi-Adeni Arabic & 38.96 & 43.29 & 57.95 & 46.74 \\
 & South Levantine Arabic & 34.78 & 51.48 & 63.29 & 49.85 \\
 & Egyptian Arabic & 40.39 & 51.02 & 63.59 & 51.67 \\
 & Tunisian Arabic & 37.02 & 58.86 & 63.01 & 52.96 \\
 & Levantine Arabic (A:North) & 38.50 & 54.75 & 68.41 & 53.89 \\
 & Moroccan Arabic & 45.98 & 56.72 & 66.55 & 56.41 \\
\cmidrule{1-6}
\multirow[t]{2}{*}{Bengali} & Standard & 45.93 & 77.52 & 57.89 & 60.45 \\
 & dhaka & 57.00 & 77.74 & 60.33 & 65.02 \\
\cmidrule{1-6}
\multirow[t]{3}{*}{Chinese} & Classical-Middle-Modern Sinitic (Traditional) & 27.67 & 25.98 & 33.66 & 29.10 \\
 & Classical-Middle-Modern Sinitic (Simplified) & 31.27 & 36.81 & 35.11 & 34.40 \\
 & Cantonese & 55.27 & 50.11 & 58.51 & 54.63 \\
\cmidrule{1-6}
\multirow[t]{3}{*}{Common turkic} & Central Oghuz & 23.86 & 37.35 & 40.88 & 34.03 \\
 & North Azerbaijani & 39.33 & 61.23 & 52.50 & 51.02 \\
 & South Azerbaijani & 54.69 & 79.78 & 91.43 & 75.30 \\
\cmidrule{1-6}
\multirow[t]{11}{*}{English} & Standard & 28.09 & 31.07 & 9.21 & 22.79 \\
 & Nigerian & 52.23 & 35.05 & 43.32 & 43.53 \\
 & Chicano & 53.75 & 38.38 & 40.75 & 44.29 \\
 & Irish & 50.63 & 39.76 & 52.41 & 47.60 \\
 & Australian & 56.82 & 39.33 & 47.42 & 47.86 \\
 & Hong Kong & 57.69 & 38.65 & 56.30 & 50.88 \\
 & Appalachian & 59.63 & 42.40 & 53.32 & 51.78 \\
 & Indian & 59.62 & 38.76 & 57.22 & 51.86 \\
 & Colloquial Singapore & 62.44 & 41.83 & 64.71 & 56.33 \\
 & African American Vernacular & 65.38 & 47.07 & 63.59 & 58.68 \\
 & Southeast American Enclave & 69.16 & 50.40 & 65.88 & 61.81 \\
\cmidrule{1-6}
\multirow[t]{24}{*}{Finnish} & Finnish & 18.91 & 43.37 & 44.81 & 35.70 \\
 & Keski-Pohjanmaa & 25.41 & 71.01 & 78.18 & 58.20 \\
 & Pohjois-Satakunta & 26.00 & 71.23 & 80.15 & 59.13 \\
 & LantinenKeski-Suomi & 21.49 & 67.64 & 88.93 & 59.35 \\
 & Etela-Satakunta & 26.84 & 71.74 & 80.08 & 59.55 \\
 & Pohjois-Hame & 23.70 & 67.54 & 88.40 & 59.88 \\
 & Etela-Pohjanmaa & 21.68 & 77.07 & 82.41 & 60.38 \\
 & Keski-Karjala & 23.46 & 75.98 & 81.75 & 60.40 \\
 & Etela-Hame & 23.19 & 69.94 & 93.25 & 62.13 \\
 & Lansipohja & 25.51 & 69.71 & 91.16 & 62.13 \\
 & Inkerinsuomalaismurteet & 22.01 & 73.31 & 91.53 & 62.28 \\
 & Kaakkois-Hame & 22.98 & 72.40 & 91.69 & 62.36 \\
 & Etela-Karjala & 23.57 & 71.89 & 91.95 & 62.47 \\
 & Pohjois-Pohjanmaa & 26.45 & 70.34 & 90.65 & 62.48 \\
 & Kainuu & 25.62 & 77.49 & 84.48 & 62.53 \\
 & Perapohjola & 24.56 & 72.24 & 91.00 & 62.60 \\
 & Pohjois-Savo & 28.17 & 77.95 & 82.29 & 62.80 \\
 & Lansi-Satakunta & 24.13 & 71.86 & 92.64 & 62.88 \\
 & Pohjois-Karjala & 28.41 & 76.42 & 84.92 & 63.25 \\
 & EtelainenKeski-Suomi & 24.57 & 72.71 & 95.50 & 64.26 \\
 & PohjoinenKeski-Suomi & 22.98 & 76.14 & 94.90 & 64.67 \\
 & Lansi-Uusimaa & 25.57 & 74.86 & 94.19 & 64.87 \\
 & Etela-Savo & 25.00 & 77.07 & 92.63 & 64.90 \\
 & PohjoinenVarsinais-Suomi & 25.86 & 76.86 & 93.94 & 65.55 \\
\cmidrule{1-6}
\multirow[t]{2}{*}{High german} & Latvian & 16.10 & 65.58 & 53.96 & 45.21 \\
 & East Latvian & 22.20 & 86.95 & 88.85 & 66.00 \\
\cmidrule{1-6}
\multirow[t]{2}{*}{Kurdish} & Northern Kurdish & 26.47 & 87.93 & 88.55 & 67.65 \\
 & Central Kurdish & 57.04 & 79.30 & 93.78 & 76.71 \\
\cmidrule{1-6}
\multirow[t]{2}{*}{Norwegian} & Norwegian Nynorsk & 17.51 & 65.52 & 45.04 & 42.69 \\
 & Norwegian Bokmal & 19.93 & 60.26 & 56.38 & 45.53 \\
\cmidrule{1-6}
\multirow[t]{2}{*}{Sotho-tswana} & Southern Sotho & 34.63 & 97.45 & 96.83 & 76.30 \\
 & Northern Sotho & 46.02 & 95.78 & 97.49 & 79.76 \\
\cmidrule{1-6}
Average &  & 35.25 & 61.56 & 70.43 & 55.75 \\
\cmidrule{1-6}
\bottomrule
\end{tabular}

    \caption{Percentage of cases where LLMs acknowledge that dialectal variations influence their judgments. The percentage is significantly lower for standard language varieties, suggesting that while LLMs recognize subtle variations in dialects, they remain robust in delivering consistent judgments.}
    \label{tab:dialect_sensitivity}
\end{table}
In this section, we delve into the LLMs' perception of dialectal sensitivity. We analyze whether the models acknowledge that dialectal variations influence their toxicity judgments. The results in \cref{tab:dialect_sensitivity} compare the LLMs' responses for both standard language datasets and their dialectal counterparts, providing insights into how dialectal nuances impact model decisions.

\subsection*{Language Variety Table}
\label{app:variety_table}
\begin{table*}
\small
\centering
\begin{tabular}{llc}
\toprule
Language Cluster & Variety Name & Example Count \\
\midrule
\multirow{9}{*}{Arabic} & North Mesopotamian Arabic & 940 \\
& Ta'izzi-Adeni Arabic & 940 \\
& Tunisian Arabic & 940 \\
& South Levantine Arabic & 940 \\
& Levantine Arabic (A:North) & 940 \\
& \underline{Standard Arabic} & 940 \\
& Najdi Arabic & 940 \\
& Moroccan Arabic & 940 \\
& Egyptian Arabic & 940 \\
\cmidrule(r){1-3}
\multirow{2}{*}{Bengali} & \underline{Standard} & 940 \\
& Dhaka Dialect & 380 \\
\cmidrule(r){1-3}
\multirow{3}{*}{Chinese} & \underline{Cantonese} & 940 \\
& Classical-Middle-Modern Sinitic (Simplified) & 940 \\
& Classical-Middle-Modern Sinitic (Traditional) & 940 \\
\cmidrule(r){1-3}
\multirow{24}{*}{Finnish} & \underline{Finnish} & 940 \\
& Pohjois-Satakunta & 940 \\
& Keski-Karjala & 940 \\
& Kainuu & 940 \\
& Etela-Pohjanmaa & 940 \\
& Etela-Satakunta & 940 \\
& Pohjois-Savo & 940 \\
& Pohjois-Karjala & 940 \\
& Keski-Pohjanmaa & 940 \\
& Kaakkois-Hame & 940 \\
& PohjoinenKeski-Suomi & 940 \\
& Pohjois-Pohjanmaa & 940 \\
& PohjoinenVarsinais-Suomi & 940 \\
& Etela-Karjala & 940 \\
& Lansi-Uusimaa & 940 \\
& Inkerinsuomalaismurteet & 940 \\
& LantinenKeski-Suomi & 940 \\
& Lansi-Satakunta & 940 \\
& Etela-Savo & 940 \\
& Lansipohja & 940 \\
& Pohjois-Hame & 940 \\
& EtelainenKeski-Suomi & 940 \\
& Etela-Hame & 940 \\
& Perapohjola & 940 \\
\cmidrule(r){1-3}
\multirow{2}{*}{Kurdish} & \underline{Central Kurdish} & 940 \\
& Northern Kurdish & 940 \\
\cmidrule(r){1-3}
\multirow{2}{*}{Norwegian} & Norwegian Nynorsk & 940 \\
& \underline{Norwegian Bokmal} & 940 \\
\cmidrule(r){1-3}
\multirow{2}{*}{Latvian} & East Latvian & 940 \\
& \underline{Latvian} & 940 \\
\cmidrule(r){1-3}
\multirow{11}{*}{English} & \underline{Standard} &  940 \\
& Southeast American Enclave &  799 \\
& Chicano &  799 \\
& Nigerian &  799 \\
& African American Vernacular &  799 \\
& Appalachian &  799 \\
& Australian &  799 \\
& Colloquial Singapore &  799 \\
& Hong Kong &  799 \\
& Indian &  799 \\
& Irish &  799 \\
\cmidrule(r){1-3}
\multirow{2}{*}{Sotho-Tswana} & \underline{Northern Sotho} & 940 \\
& Southern Sotho & 940 \\
\cmidrule(r){1-3}
\multirow{3}{*}{Common Turkic} & \underline{Central Oghuz} & 940 \\
& South Azerbaijani & 940 \\
& North Azerbaijani & 940 \\
\bottomrule
\end{tabular}
\caption{Language cluster and variety names with example count. The cluster representative that we utilize as the standard variety is underlined in each cluster.}
\label{tab:varieties}
\end{table*}

The language variety table, reported in \cref{tab:varieties}, details the specific language clusters and dialects included in our dataset. It provides an overview of the 10 language clusters and 60 varieties used in the evaluation process, along with the number of examples for each variety.

We define a \textit{language cluster} as a group consisting of a primary language and its associated dialects. Each cluster is named after the primary language, with the **cluster representative** typically being the standard form or the highest-resourced variety of that language. The remaining dialects within the cluster are referred to as the \textit{varieties} of the \textit{cluster representative}. For consistency and clarity, we follow the Glottocode naming convention~\cite{ldl-glottolog} to label the varieties, ensuring that each dialect is systematically identified.

\end{document}